\documentclass[11pt,a4paper]{article}
\usepackage[hyperref]{acl2018}
\usepackage{times}
\usepackage{latexsym}
\usepackage{graphicx}
\usepackage{mathtools}
\usepackage{url}
\usepackage{array}
\usepackage{color}

\aclfinalcopy 
\def\aclpaperid{83} 

\title{Sequence-to-Action: End-to-End Semantic Graph Generation
for Semantic Parsing}

\author{Bo Chen$^{\dagger \ddagger}$, Le Sun$^{\dagger}$, Xianpei Han$^{\dagger}$ \\
  $^{\dagger}$State Key Laboratory of Computer Science \\
  Institute of Software, Chinese Academy of Sciences, Beijing, China \\
  $^{\ddagger}$University of Chinese Academy of Sciences, Beijing, China \\
  {\tt \{chenbo,sunle,xianpei\}@iscas.ac.cn} \\}

\date{}

\begin{document}
\maketitle
\begin{abstract}
  This paper proposes a neural semantic parsing approach -- Sequence-to-Action,
  which models semantic parsing as an end-to-end semantic graph generation process.
  Our method simultaneously leverages the advantages from two recent promising
  directions of semantic parsing. Firstly, our model uses a semantic graph to
  represent the meaning of a sentence, which has a tight-coupling with knowledge bases.
  Secondly, by leveraging
  the powerful representation learning and prediction ability of neural network models, we propose
  a RNN model which can effectively map sentences to action sequences for
  semantic graph generation. Experiments show that our method achieves
  state-of-the-art performance on \textsc{Overnight} dataset and gets competitive
  performance on \textsc{Geo} and \textsc{Atis} datasets.
\end{abstract}

\section{Introduction}

Semantic parsing aims to map natural language sentences to logical forms
\cite{zelle:aaai96,DBLP:conf/uai/ZettlemoyerC05,wong-mooney:2007:ACLMain,lu-EtAl:2008:EMNLP,kwiatkowski-EtAl:2013:EMNLP}.
For example, the sentence ``\textit{Which states border Texas?}''
will be mapped to {\tt answer (A, (state (A), next\_to (A, stateid ( texas ))))}.

A semantic parser needs two functions, one for structure prediction and the other
for semantic grounding. Traditional semantic parsers are usually based on compositional
grammar, such as CCG
\cite{DBLP:conf/uai/ZettlemoyerC05,zettlemoyer-collins:2007:EMNLP-CoNLL2007},
DCS \cite{liang-jordan-klein:2011:ACL-HLT2011}, etc.
These parsers compose structure using manually designed grammars, use lexicons for semantic
grounding, and exploit features for candidate logical forms ranking.
Unfortunately, it is challenging to design grammars and learn accurate lexicons,
especially in wide-open domains. Moreover, it is often hard to design effective features,
and its learning process is not end-to-end. To resolve the above problems, two promising
lines of work have been proposed: Semantic graph-based methods and Seq2Seq methods.

\begin{figure}
  \setlength{\abovecaptionskip}{0.0cm}
  \setlength{\belowcaptionskip}{-0.4cm}
   \centering
   \includegraphics[width=0.5\textwidth]{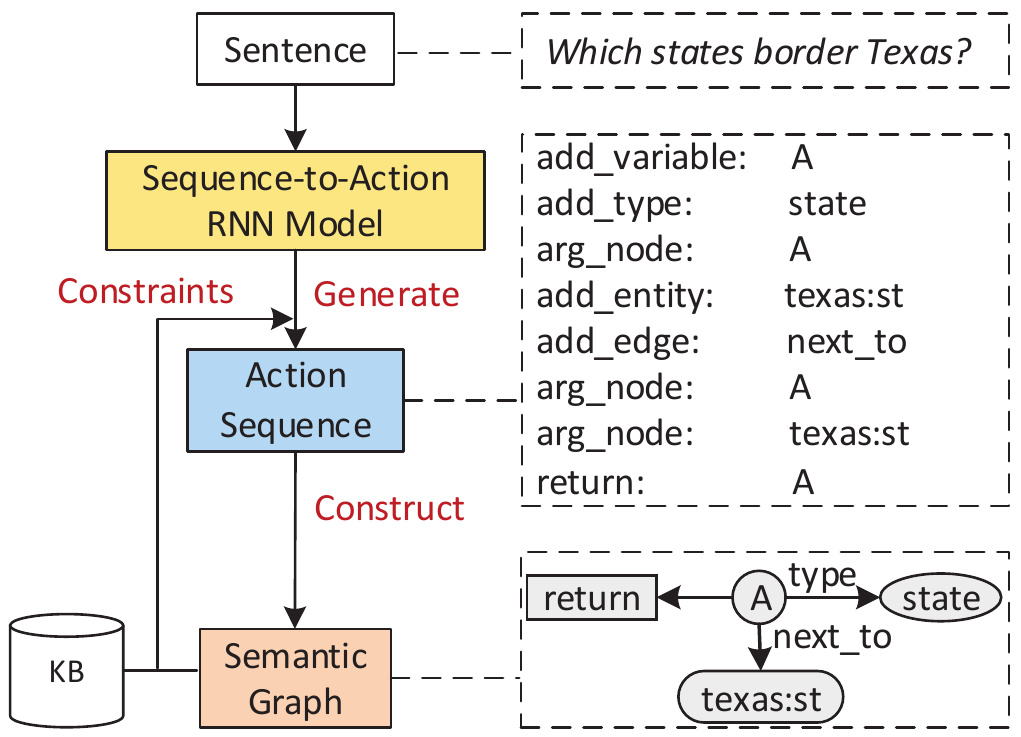}
   \caption{Overview of our method, with a demonstration example.}
   \label{figure-f0}
\end{figure}

Semantic graph-based methods
\cite{reddy_largescale_2014,reddy_transforming_2016,DBLP:conf/cikm/BastH15,yih-EtAl:2015:ACL-IJCNLP}
represent the meaning of a sentence as a semantic graph
(i.e., a sub-graph of a knowledge base, see example in Figure \ref{figure-f0}) and treat semantic parsing as a semantic
graph matching/generation process. Compared with logical
 forms, semantic graphs have a tight-coupling with knowledge bases \cite{yih-EtAl:2015:ACL-IJCNLP},
 and share many commonalities with syntactic structures \cite{reddy_largescale_2014}.
 Therefore both the structure and semantic constraints from knowledge bases can be easily
 exploited during parsing \cite{yih-EtAl:2015:ACL-IJCNLP}.
 The main challenge of semantic graph-based parsing is how to effectively construct the semantic graph
 of a sentence. Currently, semantic graphs are either constructed by matching with
 patterns \cite{DBLP:conf/cikm/BastH15}, transforming from dependency tree
 \cite{reddy_largescale_2014,reddy_transforming_2016},
 or via a staged heuristic search algorithm
 \cite{yih-EtAl:2015:ACL-IJCNLP}. These methods are all
 based on manually-designed, heuristic construction processes, making them hard to handle
 open/complex situations.

In recent years, RNN models have achieved success in sequence-to-sequence problems due to
 its strong ability on both representation learning and prediction, e.g., in machine translation
 \cite{cho-EtAl:2014:EMNLP2014}.
 A lot of Seq2Seq models have also been employed for semantic parsing
 \cite{xiao-dymetman-gardent:2016:P16-1,dong-lapata:2016:P16-1,jia-liang:2016:P16-1},
 where a sentence is parsed by translating it to linearized logical form using RNN models. There is no
 need for high-quality lexicons, manually-built grammars, and hand-crafted features.
  These models are trained end-to-end, and can leverage attention mechanism
  \cite{DBLP:journals/corr/BahdanauCB14,luong-pham-manning:2015:EMNLP}
  to learn soft alignments between sentences and logical forms.

In this paper, we propose a new neural semantic parsing framework -- Sequence-to-Action,
which can simultaneously leverage the advantages of semantic graph representation and the
strong prediction ability of Seq2Seq models. Specifically, we model semantic parsing as
an end-to-end semantic graph generation process. For example in Figure \ref{figure-f0}, our model will
parse the sentence ``\textit{Which states border Texas}'' by generating a sequence of actions
[\texttt{add\_variable:A}, \texttt{add\_type:state}, ...]. To achieve the above goal, we first design an action set
which can encode the generation process of semantic graph (including node actions such as \texttt{add\_variable},
\texttt{add\_entity}, \texttt{add\_type}, edge actions such as \texttt{add\_edge}, and operation actions such as
\texttt{argmin}, \texttt{argmax}, \texttt{count}, \texttt{sum}, etc.). And then we design a RNN model which can generate
the action sequence for constructing the semantic graph of a sentence. Finally we
further enhance parsing by
incorporating both structure and semantic constraints during decoding.

Compared with the manually-designed, heuristic
generation algorithms used in traditional semantic graph-based methods,
our sequence-to-action method generates semantic graphs
using a RNN model, which is learned end-to-end from training data.
 Such a learnable, end-to-end generation makes our approach more effective and can fit to different situations.

Compared with the previous Seq2Seq semantic parsing methods, our sequence-to-action
model predicts a sequence of semantic graph generation actions, rather than linearized
logical forms. We find that the action sequence encoding can better capture structure and semantic
 information, and is more compact. And the parsing can be enhanced by exploiting structure and semantic
 constraints.
 For example, in \textsc{Geo} dataset, the action \texttt{add\_edge:next\_to} must subject to the semantic
 constraint that its arguments must be of type \texttt{state} and \texttt{state}, and the structure
 constraint that the edge \texttt{next\_to} must connect two nodes to form a valid graph.

We evaluate our approach on three standard datasets: \textsc{Geo} \cite{zelle:aaai96},
\textsc{Atis} \cite{HE200585} and \textsc{Overnight} \cite{wang-berant-liang:2015:ACL-IJCNLP}.
The results show that our method achieves state-of-the-art performance on
 \textsc{Overnight} dataset and gets
competitive performance on \textsc{Geo} and \textsc{Atis} datasets.

The main contributions of this paper are summarized as follows:

\begin{itemize}
\item We propose a new semantic parsing framework -- Sequence-to-Action, which models
semantic parsing as an end-to-end semantic graph generation process. This new framework
can synthesize the advantages of semantic graph representation and the prediction ability
of Seq2Seq models.
\item We design a sequence-to-action model, including an action set encoding for semantic
 graph generation and a Seq2Seq RNN model for action sequence prediction. We further
 enhance the parsing by exploiting structure and semantic constraints during decoding.
Experiments validate the effectiveness of our method.
\end{itemize}

\section{Sequence-to-Action Model for End-to-End Semantic Graph Generation}
Given a sentence $X= x_{1},...,x_{|X|}$, our sequence-to-action model generates
a sequence of actions $Y= y_{1},...,y_{|Y|}$
for constructing the correct semantic graph. Figure \ref{figure-f1} shows
an example. The conditional probability $P(Y|X)$ used in our model is
decomposed as follows:

\begin{equation}
\label{equation1}
  P(Y|X)=\prod_{t=1}^{|Y|}P(y_{t}|y_{<t},X)
\end{equation}

\noindent where $y_{<t}=y_{1},...,y_{t-1}$.

To achieve the above goal, we need: 1) an action set which can encode semantic
graph generation process; 2) an encoder which encodes natural language input $X$ into
a vector representation, and a decoder which generates
$y_{1},...,y_{|Y|}$ conditioned on
the encoding vector. In following we describe them in detail.

\subsection{Actions for Semantic Graph Generation}
Generally, a semantic graph consists of nodes (including variables, entities, types)
and edges (semantic relations), with some universal operations
(e.g., \texttt{argmax}, \texttt{argmin},
\texttt{count}, \texttt{sum}, and \texttt{not}). To generate a semantic graph,
we define six types of actions as follows:

\begin{figure}
    \setlength{\abovecaptionskip}{0.0cm}
    \setlength{\belowcaptionskip}{-0.4cm}
   \centering
   \includegraphics[width=0.5\textwidth]{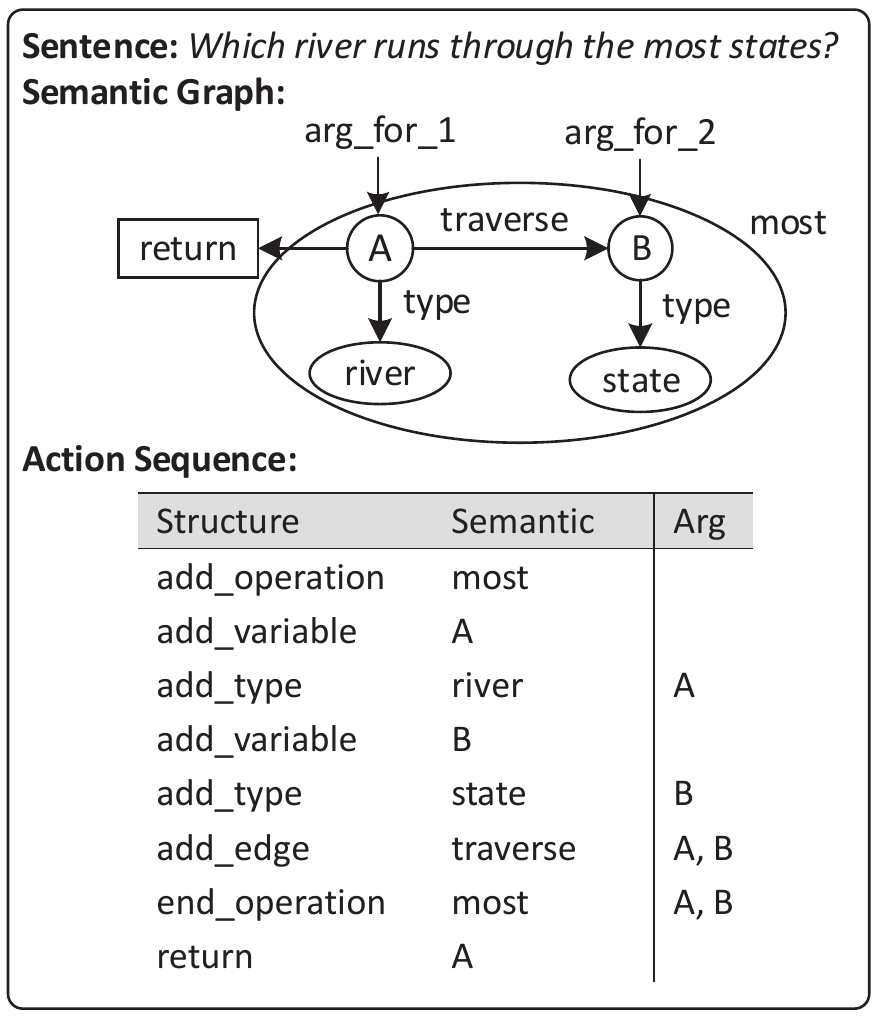}
   \caption{An example of a sentence paired with its semantic graph, together with the action
   sequence for semantic graph generation.}
   \label{figure-f1}
\end{figure}

 \textbf{Add Variable Node:} This kind of actions denotes adding a
variable node to semantic graph. In most cases a variable node is a return node
(e.g., \textit{which}, \textit{what}),
but can also be an intermediate variable node. We represent this kind of action as
\texttt{add\_variable:A}, where \texttt{A} is the identifier of the variable node.

\textbf{Add Entity Node:} This kind of actions denotes adding an entity node
(e.g., \textit{Texas}, \textit{New York}) and is represented
as \texttt{add\_entity\_node:texas}. An entity node
corresponds to an entity in knowledge bases.

\textbf{Add Type Node:} This kind of actions denotes
adding a type node (e.g., \textit{state},
\textit{city}). We represent them as \texttt{add\_type\_node:state}.

\textbf{Add Edge:} This kind of actions denotes adding an edge
between two nodes. An edge
 is a binary relation in knowledge bases. This kind of actions is represented
 as \texttt{add\_edge:next\_to}.

\textbf{Operation Action:} This kind of actions denotes adding
an operation. An operation
can be \texttt{argmax}, \texttt{argmin}, \texttt{count}, \texttt{sum}, \texttt{not}, et al.
Because each operation has a scope,
we define two actions for an operation, one is operation start action, represented as
\texttt{start\_operation:most}, and the other is operation end action, represented as
\texttt{end\_operation:most}. The subgraph within the start and end
operation actions is its scope.

\textbf{Argument Action:} Some above actions need argument information. For example,
which nodes the \texttt{add\_edge:next\_to} action should connect to. In this paper, we design argument
actions for \texttt{add\_type}, \texttt{add\_edge} and \texttt{operation} actions, and the argument actions should be
put directly after its main action.

For \texttt{add\_type} actions, we put an argument action to indicate which node this type node
should constrain. The argument can be a variable node or an entity node. An argument
action for a type node is represented as \texttt{arg:A}.

For \texttt{add\_edge} action, we use two argument actions: \texttt{arg1\_node}
and \texttt{arg2\_node}, and they are
represented as \texttt{arg1\_node:A} and \texttt{arg2\_node:B}.

We design argument actions for different operations.
For \texttt{operation:sum}, there are three
arguments: \texttt{arg-for}, \texttt{arg-in} and \texttt{arg-return}.
For \texttt{operation:count}, they are \texttt{arg-for} and
\texttt{arg-return}. There are two \texttt{arg-for} arguments for \texttt{operation:most}.

We can see that each action encodes both structure and semantic information, which makes it
easy to capture more information for parsing and can be tightly coupled with knowledge
base. Furthermore, we find that action sequence encoding is more compact than
linearized logical form (See Section \ref{section4.4}
for more details).

\subsection{Neural Sequence-to-Action Model}
\begin{figure}
   \setlength{\abovecaptionskip}{0.0cm}
   \setlength{\belowcaptionskip}{-0.2cm}
    \centering
    \includegraphics[width=0.5\textwidth]{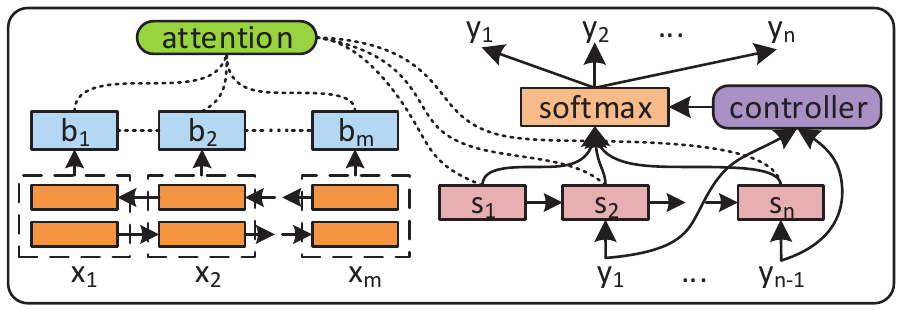}
    \caption{Our attention-based Sequence-to-Action RNN
    model, with a controller for incorporating constraints.}
    \label{figure-f2}
 \end{figure}

Based on the above action encoding mechanism, this section describes our encoder-decoder
 model for mapping sentence to action sequence.
Specifically, similar to the RNN model in
\citet{jia-liang:2016:P16-1}, this paper employs
 the attention-based sequence-to-sequence RNN model. Figure \ref{figure-f2} presents
 the overall structure.

\noindent \textbf{Encoder:} The encoder converts the input
sequence $x_{1},...,x_{m}$ to a sequence of
context-sensitive vectors $b_{1},...,b_{m}$ using a
bidirectional RNN \cite{DBLP:journals/corr/BahdanauCB14}.
Firstly each word $x_{i}$ is mapped to its embedding vector, then these vectors are fed
into a forward RNN and a backward RNN. The sequence of hidden states $h_{1},...,h_{m}$ are
generated by recurrently applying the recurrence:
 \begin{equation}
   h_{i}=LSTM(\phi^{(x)}(x_{i}),h_{i-1}).
\end{equation}
The recurrence takes the form of LSTM \cite{Hochreiter:1997:LSM:1246443.1246450}.
Finally, for each input position $i$, we define its context-sensitive embedding as
$b_{i}=[h_{i}^{F},h_{i}^{B}]$.

\noindent \textbf{Decoder:} This paper uses the classical attention-based
decoder \cite{DBLP:journals/corr/BahdanauCB14},
which generates action sequence $y_{1},...,y_{n}$, one action at a time.
At each time step $j$,
it writes $y_{j}$ based on the current hidden state $s_{j}$, then updates the hidden state
to $s_{j+1}$ based on $s_{j}$ and $y_{j}$. The decoder is formally defined by the following equations:

\begin{align}
  s_{1}&=\tanh(W^{(s)}[h_{m}^{F},h_{1}^{B}]) \\
  e_{ji}&=s_{j}^{T}W^{(a)}b_{i} \\
  a_{ji}&=\frac{\exp(e_{ji})}{{\sum_{i^{'}=1}^{m}}\exp(e_{ji^{'}})} \\
  c_{j}&=\sum_{i=1}^{m}a_{ji}b_{i} \\
  P(y_{j}&=w|x,y_{1:j-1}) \propto \exp(U_{w}[s_{j},c_{j}]) \\
  s_{j+1}&=LSTM([\phi^{(y)}(y_{j}),c_{j}],s_{j})
\end{align}

\noindent where the normalized attention scores $a_{ji}$ defines the probability distribution over input
words, indicating the attention probability on input word $i$ at time $j$;
$e_{ji}$ is un-normalized attention score. To incorporate constraints
during decoding, an extra controller component is added and its details will be described in Section \ref{section3.3}.

\noindent \textbf{Action Embedding.} The above decoder needs the embedding of each action. As described
above, each action has two parts, one for structure
(e.g., \texttt{add\_edge}), and the other for
semantic (e.g., \texttt{next\_to}). As a result, actions may share the same structure or semantic
part, e.g., \texttt{add\_edge:next\_to} and \texttt{add\_edge:loc} have the same structure part,
and \texttt{add\_node:A} and \texttt{arg\_node:A} have the same semantic part. To make parameters more compact,
we first embed the structure part and the semantic part independently, then concatenate
them to get the final embedding. For instance,
$\phi^{(y)}($\texttt{add\_edge:next\_to} $)=[$ $\phi_{strut}^{(y)}($ \texttt{add\_edge} $),\phi_{sem}^{(y)}($ \texttt{next\_to} $)]$.
The action embeddings $\phi^{(y)}$ are learned during training.

\section{Constrained Semantic Parsing using Sequence-to-Action Model}
In this section, we describe how to build a neural semantic parser using
sequence-to-action model. We first describe the training and the inference
of our model, and then introduce how to incorporate structure and semantic
constraints during decoding.

\subsection{Training}
\noindent \textbf{Parameter Estimation.} The parameters of our model
include RNN parameters $W^{(s)}$, $W^{(a)}$, $U_{w}$, word embeddings $\phi^{(x)}$,
and action embeddings $\phi^{(y)}$.  We estimate these
 parameters from training data. Given a training example with a
 sentence $X$ and its action sequence $Y$, we maximize the likelihood of the generated
 sequence of actions given $X$. The objective function is:

 \begin{equation}
    \sum_{i=1}^{n}\log P(Y_{i}|X_{i})
 \end{equation}

\noindent Standard stochastic gradient descent algorithm is
employed to update parameters.

\begin{figure}[t!]
  \setlength{\abovecaptionskip}{0.0cm}
  \setlength{\belowcaptionskip}{-0.4cm}
   \centering
   \includegraphics[width=0.5\textwidth]{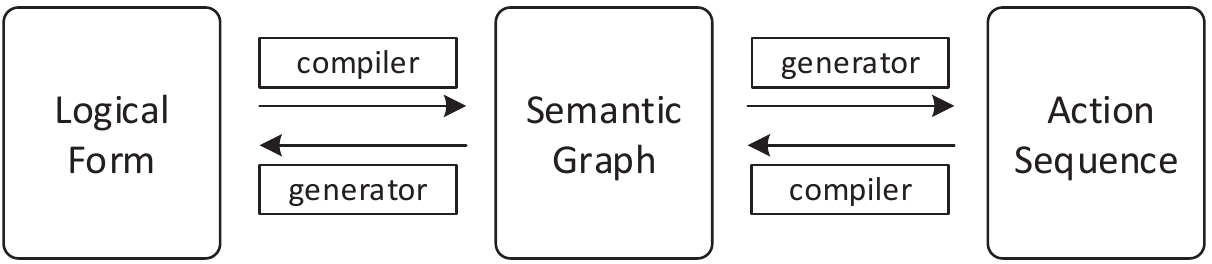}
   \caption{The procedure of converting
   between logical form and action sequence.}
   \label{figure-f3}
\end{figure}

\noindent \textbf{Logical Form to Action Sequence.} Currently, most datasets of
semantic parsing are labeled with logical forms. In order to train our model, we convert
logical forms to action sequences using semantic
graph as an intermediate representation (See Figure \ref{figure-f3} for an overview).
Concretely, we transform logical forms into semantic graphs using a depth-first-search algorithm from root,
and then generate the action sequence using the same order.
Specifically, entities, variables and types are nodes; relations are edges.
Conversely we can convert action sequence to logical form similarly.
Based on the above algorithm,
action sequences can be transformed into logical forms in a deterministic way, and the same for logical forms to action sequences.

\noindent \textbf{Mechanisms for Handling Entities.} Entities play an important role in
semantic parsing \cite{yih-EtAl:2015:ACL-IJCNLP}.
In \citet{dong-lapata:2016:P16-1}, entities are replaced
with their types and unique IDs.
In \citet{jia-liang:2016:P16-1}, entities are generated via
attention-based copying mechanism helped with a lexicon. This paper implements
both mechanisms and compares them in experiments.

\subsection{Inference}
Given a new sentence $X$, we predict action sequence by:

\begin{equation}
    Y^{*}=\underset{Y}{\operatorname{argmax}} P(Y|X)
\end{equation}

\noindent where $Y$ represents action sequence, and $P(Y|X)$ is computed using Formula \eqref{equation1}.
Beam search is used for best action sequence decoding.
Semantic graph and logical form can be derived from $Y^{*}$ as described in above.

\subsection{Incorporating Constraints in Decoding}
\label{section3.3}
For decoding, we generate action sequentially. It is obviously that the next action
has a strong correlation with the partial semantic graph generated to current,
and illegal actions can be filtered using structure and semantic constraints.
Specifically, we incorporate constraints in decoding using a controller.
This procedure has two steps: 1) the controller constructs partial semantic graph using the actions generated to current;
2) the controller checks whether a new generated action can meet all structure/semantic constraints using the partial semantic graph.

\begin{figure}
  \setlength{\abovecaptionskip}{0.0cm}
  \setlength{\belowcaptionskip}{-0.4cm}
   \centering
   \includegraphics[width=0.5\textwidth]{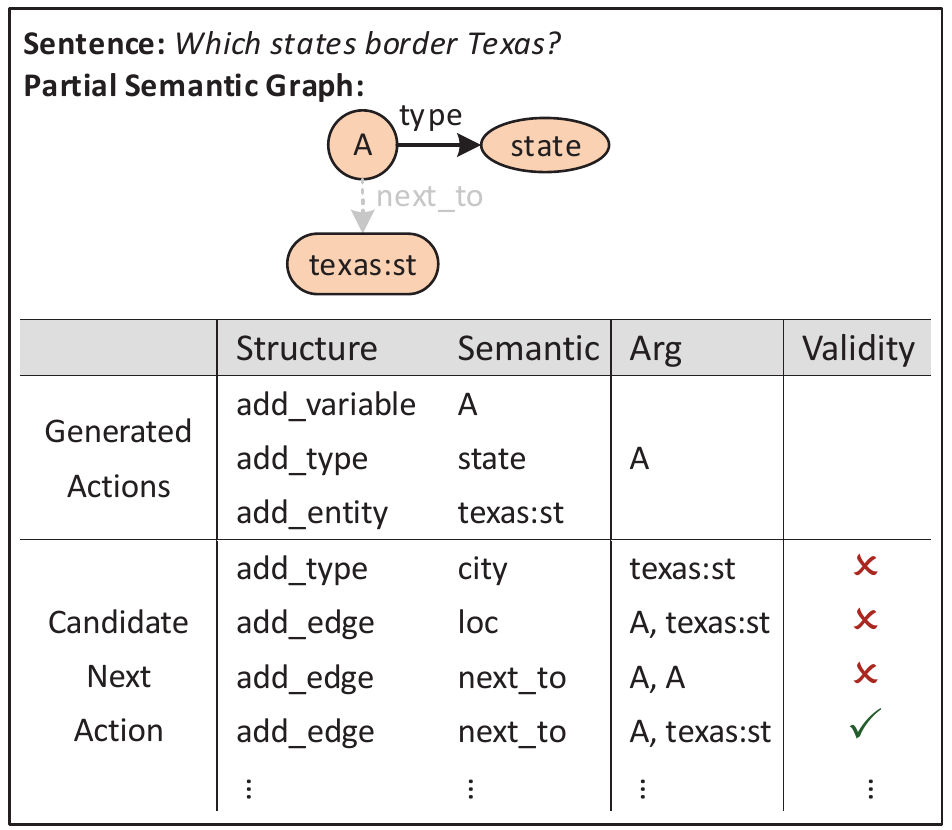}
   \caption{A demonstration of illegal action filtering using constraints.
   The graph in color is the constructed semantic graph to current.}
   \label{figure-f4}
\end{figure}

\noindent \textbf{Structure Constraints.} The structure constraints ensure action sequence
will form a connected acyclic graph. For example, there must be two argument nodes for
an edge, and the two argument nodes should be different
(The third candidate next action in Figure \ref{figure-f4} violates this constraint).
This kind of constraints are
domain-independent. The controller encodes structure constraints as a set of rules.

\noindent \textbf{Semantic Constraints.} The semantic constraints ensure the constructed
graph must follow the schema of knowledge bases. Specifically, we model two types of semantic
constraints. One is selectional preference constraints where the argument types of a
relation should follow knowledge base schemas. For example, in \textsc{Geo} dataset,
relation \texttt{next\_to}'s \texttt{arg1}
and \texttt{arg2} should both be a \texttt{state}. The second is type conflict constraints,
i.e., an entity/variable node's type must be consistent, i.e., a node cannot be both
of type \texttt{city} and \texttt{state}. Semantic constraints are domain-specific and are automatically
extracted from knowledge base schemas. The controller encodes semantic constraints as a
set of rules.

\section{Experiments}
In this section, we assess the performance of our method
and compare it with previous methods.

\subsection{Datasets}
We conduct experiments on three standard datasets: \textsc{Geo}, \textsc{Atis} and \textsc{Overnight}.

\noindent \textbf{\textsc{Geo}} contains natural language questions about US geography paired
with corresponding Prolog database queries.
Following \citet{DBLP:conf/uai/ZettlemoyerC05},
we use the standard 600/280 instance splits for training/test.

\noindent \textbf{\textsc{Atis}} contains natural language questions of a flight database,
with each question is annotated with a lambda calculus query.
Following \citet{zettlemoyer-collins:2007:EMNLP-CoNLL2007}, we use the standard 4473/448 instance
splits for training/test.

\noindent \textbf{\textsc{Overnight}} contains natural language paraphrases paired with
logical forms across eight domains.
We evaluate on the standard train/test splits
as \citet{wang-berant-liang:2015:ACL-IJCNLP}.

\subsection{Experimental Settings}
Following the experimental setup of \citet{jia-liang:2016:P16-1}: we use 200 hidden units and
100-dimensional word vectors for sentence encoding. The dimensions of action embedding
are tuned on validation datasets for each corpus. We initialize all parameters by uniformly
sampling within the interval [-0.1, 0.1]. We train our model for a total of 30 epochs
with an initial learning rate of 0.1, and halve the learning rate every 5 epochs after
epoch 15. We replace word vectors for words occurring only once with an universal
word vector. The beam size is set as 5. Our model is implemented in
Theano \cite{bergstra2010theano}, and the codes and settings are released on Github:
https://github.com/dongpobeyond/Seq2Act.

We evaluate different systems using the standard accuracy metric, and the accuracies on
different datasets are obtained as same as \citet{jia-liang:2016:P16-1}.

\subsection{Overall Results}
We compare our method with state-of-the-art systems on all three datasets.
Because all systems using the same training/test splits,
we directly use the reported best performances from their original papers for fair comparison.

For our method, we train our model with three settings: the first one is the basic
sequence-to-action model without constraints -- Seq2Act; the second one adds structure
constraints in decoding -- Seq2Act (+C1); the third one is the full model which adds both structure and
semantic constraints -- Seq2Act (+C1+C2).
Semantic constraints (C2) are stricter than structure constraints (C1). Therefore we set that C1 should be first met for C2 to be met.
So in our experiments we add constraints incrementally.
The overall results are shown in Table \ref{table1}-\ref{table2}. From the overall results, we can see that:

\begin{table}[t!]
  \setlength{\abovecaptionskip}{0.0cm}
  \setlength{\belowcaptionskip}{-0.4cm}
\begin{center}
\begin{tabular}{|l|c|c|}
\hline  & \textsc{Geo} &  \textsc{Atis} \\ \hline
\textbf{Previous Work} &   &  \\
\citet{DBLP:conf/uai/ZettlemoyerC05} & 79.3 & -- \\
\citet{zettlemoyer-collins:2007:EMNLP-CoNLL2007} & 86.1 & 84.6 \\
\citet{kwiatkowksi-EtAl:2010:EMNLP} & 88.9 & -- \\
\citet{kwiatkowski-EtAl:2011:EMNLP} & 88.6 & 82.8 \\
\citet{liang-jordan-klein:2011:ACL-HLT2011}* (+lexicon) & \textbf{91.1} & -- \\
\citet{poon:2013:ACL2013} & -- & 83.5 \\
\citet{zhao-hassan-auli:2015:NAACL-HLT} & 88.9 & 84.2 \\
\citet{rabinovich-stern-klein:2017:Long} & 87.1 & \textbf{85.9} \\\hline
\textbf{Seq2Seq Models} &  & \\
\citet{jia-liang:2016:P16-1} & 85.0 & 76.3\\
\citet{jia-liang:2016:P16-1}* (+data) & 89.3 & 83.3 \\
\citet{dong-lapata:2016:P16-1}: 2Seq & 84.6 & 84.2 \\
\citet{dong-lapata:2016:P16-1}: 2Tree & 87.1 & 84.6 \\  \hline
\textbf{Our Models} &  & \\
Seq2Act & 87.5 & 84.6 \\
Seq2Act (+C1) & 88.2 & 85.0 \\
Seq2Act (+C1+C2) & 88.9 & 85.5 \\
\hline
\end{tabular}
\end{center}
\caption{\label{table1} Test accuracies on
                        \textsc{Geo} and \textsc{Atis} datasets, where * indicates systems
                        with extra-resources are used.}
\end{table}

\begin{table*}[t!]
  \setlength{\abovecaptionskip}{0.0cm}
  \setlength{\belowcaptionskip}{-0.1cm}
\begin{center}
\begin{tabular}{|l|cccccccc|c|}
\hline  & Soc. & Blo. & Bas. & Res. & Cal. & Hou. & Pub. & Rec. & Avg. \\ \hline
\textbf{Previous Work} & & & & & & & & & \\
\citet{wang-berant-liang:2015:ACL-IJCNLP} &
48.2 & 41.9 & 46.3 & 75.9 & 74.4 & 54.0 & 59.0 & 70.8 & 58.8 \\ \hline
\textbf{Seq2Seq Models} & & & & & & & & & \\
\citet{xiao-dymetman-gardent:2016:P16-1} &
80.0 & 55.6 & 80.5 & 80.1 & 75.0 & 61.9 & 75.8 & -- & 72.7 \\
\citet{jia-liang:2016:P16-1} &
81.4 & 58.1 & 85.2 & 76.2 & 78.0 & 71.4 & 76.4 & 79.6 & 75.8 \\
 \citet{jia-liang:2016:P16-1}* (+data) &
79.6 & 60.2 & 87.5 & 79.5 & 81.0 & 72.5 & 78.3 & 81.0 & 77.5 \\ \hline
\textbf{Our Models} & & & & & & & & & \\
Seq2Act & 81.4 & 60.4 & 87.5 & 79.8 & 81.0 & 73.0 & 79.5 & 81.5 & 78.0 \\
Seq2Act (+C1) & 81.8 & 60.9 & 88.0 & 80.1 & 81.0 & 73.5 & 80.1 & 82.0 & 78.4 \\
Seq2Act (+C1+C2) & \textbf{82.1} & \textbf{61.4} & \textbf{88.2} & \textbf{80.7}
& \textbf{81.5} & \textbf{74.1} & \textbf{80.7} & \textbf{82.9} & \textbf{79.0} \\
\hline
\end{tabular}
\end{center}
\caption{\label{table2} Test accuracies on \textsc{Overnight} dataset, which includes eight
domains: Social, Blocks, Basketball, Restaurants, Calendar, Housing, Publications, and Recipes.}
\end{table*}

1) By synthetizing the advantages of semantic graph representation and the prediction
ability of Seq2Seq model, our method achieves state-of-the-art performance on
\textsc{Overnight} dataset, and gets competitive performance on \textsc{Geo} and \textsc{Atis} dataset. In fact,
on \textsc{Geo} our full model (Seq2Act+C1+C2) also gets the best test accuracy of 88.9 if under the
same settings, which only falls behind
\citet{liang-jordan-klein:2011:ACL-HLT2011}* which uses extra hand-crafted
 lexicons and \citet{jia-liang:2016:P16-1}* which uses extra augmented training data.
 On \textsc{Atis} our full model gets the second best test accuracy of 85.5, which only falls
 behind \citet{rabinovich-stern-klein:2017:Long} which uses a supervised attention strategy.
  On \textsc{Overnight}, our full model gets state-of-the-art accuracy of 79.0,
 which even outperforms \citet{jia-liang:2016:P16-1}* with extra augmented training data.

2) Compared with the linearized logical form representation used in previous Seq2Seq
baselines, our action sequence encoding is more effective for semantic parsing.
On all three datasets, our basic Seq2Act model gets better results than all
Seq2Seq baselines. On \textsc{Geo}, the Seq2Act model achieve test accuracy of 87.5,
better than the best accuracy 87.1 of Seq2Seq baseline. On \textsc{Atis}, the Seq2Act model
obtains a test accuracy of 84.6, the same as the best Seq2Seq baseline. On \textsc{Overngiht},
the Seq2Act model gets a test accuracy of 78.0, better than the best Seq2Seq
baseline gets 77.5. We argue that this is because
our action sequence encoding is more compact and can capture more information.

3) Structure constraints can enhance semantic parsing by ensuring the
validity of graph using the generated action sequence. In all three datasets, Seq2Act (+C1)
outperforms the basic Seq2Act model. This is because a part of illegal
actions will be filtered during decoding.

4) By leveraging knowledge base schemas during decoding, semantic constraints are
effective for semantic parsing. Compared to Seq2Act and Seq2Act (+C1),
the Seq2Act (+C1+C2) gets the best performance on all three datasets. This
is because semantic constraints can further filter semantic illegal actions using
selectional preference and consistency between types.

\subsection{Detailed Analysis}
\label{section4.4}
\noindent \textbf{Effect of Entity Handling Mechanisms.} This paper implements two entity handling
mechanisms -- \textit{Replacing} \cite{dong-lapata:2016:P16-1}
which identifies entities and then replaces them with their types and IDs, and attention-based
\textit{Copying} \cite{jia-liang:2016:P16-1}. To compare the above two mechanisms,
we train and test with our full model and the results are shown in Table \ref{table3}. We can see that,
\textit{Replacing} mechanism outperforms \textit{Copying} in all three datasets.
This is because \textit{Replacing} is done in preprocessing, while attention-based \textit{Copying} is done
during parsing and needs additional copy mechanism.

\begin{table}[t!]
  \setlength{\abovecaptionskip}{0.0cm}
  \setlength{\belowcaptionskip}{-0.2cm}
\begin{center}
\begin{tabular}{|c|c|c|}
\hline   & \textit{Replacing}  & \textit{Copying}  \\ \hline
 \textsc{Geo} & 88.9 & 88.2 \\ \hline
 \textsc{Atis}& 85.5 & 84.0 \\ \hline
 \textsc{Overnight}  & 79.0 & 77.9 \\ \hline
\end{tabular}
\end{center}
\caption{\label{table3} Test accuracies of Seq2Act (+C1+C2) on \textsc{Geo}, \textsc{Atis},
and \textsc{Overnight} of two entity handling mechanisms. }
\end{table}

\noindent \textbf{Linearized Logical Form vs. Action Sequence.}
Table \ref{table4} shows the average length of linearized
logical forms used in previous
Seq2Seq models and the action sequences of our model on all three datasets. As we can see,
 action sequence encoding is more compact than linearized logical form encoding:
 action sequence is shorter on all three datasets, 35.5\%, 9.2\% and 28.5\%
 reduction in length respectively. The main advantage of a shorter/compact
 encoding is that it will reduce the influence of long distance dependency problem.

 \begin{table}[t!]
   \setlength{\abovecaptionskip}{0.0cm}
   \setlength{\belowcaptionskip}{-0.2cm}
 \begin{center}
 \begin{tabular}{|c|c|c|c|}
 \hline  &  Logical Form & Action Sequence \\ \hline
 \textsc{Geo} & 28.2 & 18.2 \\ \hline
 \textsc{Atis} & 28.4 & 25.8 \\ \hline
 \textsc{Overnight} & 46.6 & 33.3 \\ \hline
 \end{tabular}
 \end{center}
 \caption{\label{table4} Average length of logical forms and action sequences on
 three datasets. On \textsc{Overnight}, we average across all eight domains.}
 \end{table}

\subsection{Error Analysis}
We perform error analysis on results and find there are mainly two
types of errors.

\begin{table*}[t!]
  \setlength{\abovecaptionskip}{0.0cm}
  \setlength{\belowcaptionskip}{-0.2cm}
\small
\begin{center}
\begin{tabular}{|m{1.5cm}|m{13.5cm}|}
\hline  \multicolumn{1}{|c|}{Error Types} &  \multicolumn{1}{c|}{Examples}  \\ \hline
Un-covered Sentence Structure & Sentence: \textit{Iowa borders how many states?}
{\color{red}(Formal Form: \textit{How many states does Iowa border?})}

Gold Parse: \texttt{answer(A, count(B, (const (C, stateid(iowa)), next\_to(C, B), state (B)), A))}

Predicted Parse: \texttt{answer (A, count(B, state(B), A))}
 \\ \hline
Under-Mapping & Sentence: \textit{Please show me \textbf{{\color{red}first class}} flights from indianapolis to memphis one way leaving before 10am}

Gold Parse: \texttt{(lambda x (and (flight x) (oneway x) \textbf{{\color{red}(class\_type x first:cl)}} (< (departure\_time x) 1000:ti) (from x indianapolis:ci) (to x memphis:ci)))}

Predicted Parse: \texttt{(lambda x (and (flight x) (oneway x) (< (departure\_time x) 1000:ti) (from x indianapolis:ci) (to x memphis:ci)))}
 \\ \hline
\end{tabular}
\end{center}
\caption{\label{table5} Some examples for error analysis. Each example includes the
sentence for parsing, with gold parse and predicted parse from our model.}
\end{table*}

\noindent \textbf{Unseen/Informal Sentence Structure.} Some test sentences
have unseen syntactic structures.
For example, the first case in Table \ref{table5} has an unseen and informal structure,
where entity word ``\textit{Iowa}'' and relation word ``\textit{borders}'' appear
ahead of the question words ``\textit{how many}''.
 For this problem, we can employ sentence rewriting or
 paraphrasing techniques \cite{chen-EtAl:2016:P16-12,dong-EtAl:2017:EMNLP2017}
 to transform unseen sentence structures into normal ones.

\noindent \textbf{Under-Mapping.} As \citet{dong-lapata:2016:P16-1} discussed,
the attention model does not take the alignment history into consideration,
makes some words are ignored during parsing. For example in the second case in
Table \ref{table5}, ``\textit{first class}'' is ignored during the decoding process.
This problem can be further solved
using explicit word coverage models used in
neural machine translation \cite{tu-EtAl:2016:P16-1,cohn-EtAl:2016:N16-1}

\section{Related Work}

Semantic parsing has received significant attention for a long time
\cite{kate-mooney:2006:COLACL,clarke-EtAl:2010:CONLL,krishnamurthy-mitchell:2012:EMNLP-CoNLL,artzi-zettlemoyer:2013:TACL,berant-liang:2014:P14-1,quirk-mooney-galley:2015:ACL-IJCNLP,artzi-lee-zettlemoyer:2015:EMNLP,reddy-EtAl:2017:EMNLP2017,C18-1076}.
Traditional methods are mostly based on the principle of compositional semantics,
which first trigger predicates using lexicons and then compose them using grammars.
The prominent grammars include
SCFG \cite{wong-mooney:2007:ACLMain,li-EtAl:2015:EMNLP3},
CCG \cite{DBLP:conf/uai/ZettlemoyerC05,kwiatkowski-EtAl:2011:EMNLP,cai-yates:2013:ACL2013},
DCS \cite{liang-jordan-klein:2011:ACL-HLT2011,berant-EtAl:2013:EMNLP}, etc.
As discussed above, the main drawback of grammar-based methods is that they rely on
high-quality lexicons, manually-built grammars, and hand-crafted features.

In recent years, one promising direction of semantic parsing is to use semantic graph
as representation. Thus semantic parsing is modeled as a semantic graph generation process.
\citet{ge-mooney:2009:ACLIJCNLP} build semantic graph by transforming syntactic tree.
\citet{DBLP:conf/cikm/BastH15} identify the structure of a semantic query using three
pre-defined patterns.
\citet{reddy_largescale_2014,reddy_transforming_2016} use Freebase-based semantic graph
representation, and convert sentences to semantic graphs using CCG or dependency tree.
\citet{yih-EtAl:2015:ACL-IJCNLP} generate semantic graphs using a staged heuristic search algorithm.
These methods are all based on manually-designed, heuristic generation process,
which may suffer from syntactic parse errors
\cite{ge-mooney:2009:ACLIJCNLP,reddy_largescale_2014,reddy_transforming_2016},
structure mismatch \cite{chen-EtAl:2016:P16-12}, and are hard to deal with complex
sentences \cite{yih-EtAl:2015:ACL-IJCNLP}.

One other direction is to employ neural Seq2Seq models, which models semantic parsing
as an end-to-end, sentence to logical form machine translation problem.
\citet{dong-lapata:2016:P16-1},
\citet{jia-liang:2016:P16-1}
and \citet{xiao-dymetman-gardent:2016:P16-1} transform word
sequence to linearized logical forms. One main drawback of these methods is that it is
hard to capture and exploit structure and semantic constraints using linearized logical
forms.
\citet{dong-lapata:2016:P16-1} propose a Seq2Tree model to capture the hierarchical structure
of logical forms.

It has been shown that structure and semantic constraints are effective for enhancing
semantic parsing.
\citet{krishnamurthy-dasigi-gardner:2017:EMNLP2017} use type constraints to filter illegal tokens.
\citet{liang-EtAl:2017:Long} adopt a Lisp interpreter with pre-defined functions to produce
valid tokens.
\citet{iyyer-yih-chang:2017:Long} adopt type constraints to generate valid actions. Inspired by these
approaches, we also incorporate both structure and semantic constraints in our neural
sequence-to-action model.

Transition-based approaches are important in both dependency parsing \cite{Nivre:2008:ADI:1479202.1479205,Henderson:2013:MJP:2576217.2576223}
and AMR parsing \cite{wang-xue-pradhan:2015:NAACL-HLT}. In semantic parsing, our method has a tight-coupling with knowledge bases,
and constraints can be exploited for more accurate decoding. We believe this can also be used to enhance previous transition based methods and
may also be used in other parsing tasks, e.g., AMR parsing.

\section{Conclusions}

This paper proposes Sequence-to-Action, a method which models semantic parsing as
an end-to-end semantic graph generation process. By leveraging the advantages of
semantic graph representation
and exploiting the representation learning and prediction ability of Seq2Seq models, our method
achieved significant performance improvements on three datasets.
Furthermore, structure and semantic constraints can be easily incorporated in decoding
to enhance semantic parsing.

For future work, to solve the problem of the lack of training data, we want to design weakly
supervised learning algorithm using denotations (QA pairs) as supervision.
Furthermore, we want to collect labeled data by designing an interactive UI
for annotation assist like \cite{yih-EtAl:2016:P16-2}, which uses semantic graphs to annotate the
meaning of sentences, since semantic graph is more natural and can be easily
annotated without the need of expert knowledge.

\bibliography{acl2018}
\bibliographystyle{acl_natbib}

\end{document}